\documentclass[journal,twoside]{IEEEtran}

\usepackage{cite}
\usepackage{times}
\usepackage{epsfig}
\usepackage{graphicx}
\usepackage{algorithm}
\usepackage{algorithmic}
\usepackage{comment}
\usepackage{textcomp}
\usepackage{amsmath,amsfonts,amssymb}
\usepackage{pifont}
\usepackage{booktabs}
\usepackage{bm}
\usepackage{multirow}
\usepackage{multicol}
\usepackage{enumitem}
\usepackage{makecell}
\usepackage{array}
\usepackage[caption=false,font=normalsize,labelfont=sf,textfont=sf]{subfig}
\usepackage{xcolor}
\usepackage{isomath}
\usepackage{stmaryrd}
\usepackage{tikz}
\usepackage{hyperref}
\usepackage{lipsum}
\usepackage{orcidlink}
\usepackage{soul}

\newcommand{\xmark}{\ding{55}}
\newcommand{\mytabheader}[1]{\textbf{\textit{#1}}}
\renewcommand{\matrixsym}[1]{\bm{#1}}

\newcommand{\bspec}{$\text{B}_{1,8A,11,12}$}

\begin{document}
%
\title{Spotting Virus from Satellites: Modeling the Circulation of West Nile Virus Through Graph Neural Networks}

\author{Lorenzo~Bonicelli\orcidlink{0000-0002-9717-5602},
Angelo~Porrello\orcidlink{0000-0002-9022-8484},
Stefano~Vincenzi\orcidlink{0000-0003-4401-0094},
Carla~Ippoliti\orcidlink{0000-0001-9030-7329},
Federica~Iapaolo\orcidlink{0000-0002-3627-915X},
Annamaria~Conte\orcidlink{0000-0001-6975-2698},
Simone~Calderara\orcidlink{0000-0001-9056-1538},~\IEEEmembership{Member,~IEEE,}%
\thanks{L. Bonicelli, A. Porrello, S. Vincenzi, and S. Calderara are with University of Modena and Reggio Emilia, Modena, Italy.}
\thanks{C. Ippoliti, F. Iapaolo, and A. Conte are with Istituto Zooprofilattico Sperimentale dell’Abruzzo e del Molise ‘G.Caporale’, Teramo, Italy.}
}


%
%

\markboth{}%
{Bonicelli \MakeLowercase{\textit{et al.}}: Spotting Virus from Satellites}
%



\maketitle

\begin{abstract}
    The occurrence of West Nile Virus (WNV) represents one of the most common mosquito-borne zoonosis viral infections. Its circulation is usually associated with climatic and environmental conditions suitable for vector proliferation and virus replication. On top of that, several statistical models have been developed to shape and forecast WNV circulation: in particular, the recent massive availability of Earth Observation (EO) data, coupled with the continuous advances in the field of Artificial Intelligence, offer valuable opportunities.
    
    In this paper, we seek to predict WNV circulation by feeding Deep Neural Networks (DNNs) with satellite images, which have been extensively shown to hold environmental and climatic features. Notably, while previous approaches analyze each geographical site independently, we propose a spatial-aware approach that considers also the characteristics of close sites. Specifically, we build upon Graph Neural Networks (GNN) to aggregate features from neighbouring places, and further extend these modules to consider multiple relations, such as the difference in temperature and soil moisture between two sites, as well as the geographical distance. Moreover, we inject time-related information directly into the model to take into account the seasonality of virus spread.
    
    We design an experimental setting that combines satellite images -- from Landsat and Sentinel missions -- with ground truth observations of WNV circulation in Italy. We show that our proposed Multi-Adjacency Graph Attention Network (MAGAT) consistently leads to higher performance when paired with an appropriate pre-training stage. Finally, we assess the importance of each component of MAGAT in our ablation studies.
\end{abstract}


\section{Introduction}
Zoonoses\footnote{This work has been submitted to the IEEE Transactions On Geoscience And Remote Sensing for possible publication. Copyright may be transferred without notice, after which this version may no longer be accessible.} are diseases that are transmissible from animals to humans. Depending on the way of transmission, they can be foodborne, waterborne, vector-borne, transmitted through direct contact with animals, or indirectly by fomites or environmental contamination. These diseases represent a severe threat to worldwide public health by now, constituting approximately 60\% of all emerging infectious diseases reported globally~\cite{taylor2001risk}. For this reason, considerable efforts have recently been made to set up integrated surveillance plans~\cite{glews,fao2019taking} paving the way towards early recognition and intervention of critical settings.

In such a scenario, West Nile Virus (WNV) infection is one of the most widespread zoonosis in Eastern, Western and Southern Europe. The disease is caused by the West Nile virus -- a positive-strand RNA flavivirus -- and is commonly transmitted by the bite of an infected mosquito. Incidentally, the transmission cycle can lead to human infections, where around 1 in 5 people present flu-like symptoms and 1 in 150 may develop a more serious -- sometimes even fatal -- illness. Several bird species are the main hosts of WNV~\cite{komar2003experimental,rizzoli2015understanding,spedicato2016experimental,mencattelli2022west}, but hundreds of cases each year concern infections in humans and other mammals (\textit{e.g.}, horses) considered dead-end hosts. Although most of these cases are asymptomatic, the evidence of viral circulation can be associated with clinical symptoms. 

The persistence of the virus in nature is favoured by various elements: the vector presence as mentioned, but also climatic and environmental factors play an important role~\cite{paz2013environmental,kilpatrick2008temperature}. For example, the transmission of the pathogen is highly influenced by temperature, which determines both the survival conditions~\cite{jetten1997potential} and the intensity (\textit{i.e.}, the fractions of vectors that carry the disease) of virus spread~\cite{dohm2002effect}. Incidentally, climate change has been implicated as a contributing cause for the changing patterns of WNV transmission~\cite{paz2006west,semenza2009climate,tabachnick2010challenges,ebi2013adaptation,charles2016impacts,semenza2018vector,semenza2021climate}.

On top of this evidence, previous works~\cite{chevalier2014predictive,tran2014environmental,marcantonio2015identifying,conte2015spatio} attempted to model and track the spread of WNV infection by examining climatic-environmental variables in an automated fashion. They typically rely on Earth Observation (EO) to extract indicators that describe the land cover, and are based on Land Surface Temperature (LST), Normalized Difference Vegetation Index (NDVI), and Modified Normalised Difference Water Index (MNDWI). Even if these derived indices have led to satisfactory results, the growth and massive availability of new satellite data from more recent space missions have enabled the exploitation of Deep Learning (DL) techniques for agriculture~\cite{zhang2017band}, or insect population models~\cite{vincenzi2019spotting}. Concerning those problems dealing with animal and human health, the DL paradigm appears promising for modeling the environmental and climate factors, as these methodologies have the potential to learn the most appropriate high-level features directly from raw spectral band satellite images. Notably, these techniques took a few years to emerge as state-of-the-art for diverse remote sensing tasks, including soil and crop classification~\cite{zhang2017band}, image fusion~\cite{yu2018deep} and change detection~\cite{gong2016change,shafique2022deep}.

Following this trend, our research interest regards the analysis of the spread of the West Nile disease directly from remote sensing data. Specifically, we cast the problem as binary classification and propose a novel method that extracts both temporal and spatial cues from satellite imagery. In a slightly similar vein, recent works~\cite{vincenzi2019spotting,vincenzi2021color} have tackled the issue of detecting not the circulation of the virus, but instead the presence/absence of the corresponding vector. These approaches usually examine the satellite imagery centered on a certain geographical site and attempt to infer the probability of vector presence. However, as a single point is considered for inference, the estimation provided by the model could be susceptible to outliers, obstructions (\textit{e.g.}, clouds, fog), noisy acquisitions, or other artifacts.

With the aim of obtaining robust predictions, our approach resorts to taking into account also the relations between geographical sites: namely, areas with similar climatic and environmental conditions are potentially exposed to similar disease risks, even if they are geographically distant from each other~\cite{ippoliti2019defining}. On top of that relation, we mainly contribute to the field by extending the input of the model and including a neighbourhood of spatially-close samples around the point of interest, thus studying the circulation at a wider geographical scale.

In methodological terms, we build upon the work of~\cite{vincenzi2021color} and extract -- for each site independently -- a set of high-level features from raw multi-band input. In addition, we aid the feature extractor to catch the seasonal patterns peculiar to the WNV circulation: we do so by conditioning~\cite{courville2017modulating} the computations of the internal representations on the month that spectral bands were captured by sensors. Then, to encompass the environmental and geographical relations between nearby sites, we arrange their high-level representations as the nodes of a graph spanning multiple geographical locations. Thanks to such a formulation, we can model the edge between two sites by considering not only their geographical distance (\textit{e.g.,} using the Haversine formula) but also their affinity in terms of environmental and climatic variables, such as temperature and soil moisture. We then process the graph through a modified version of Graph Attention Networks (GAT)~\cite{velivckovic2017graph}, revised to deal with multiple real-valued adjacency matrices. With the adoption of this architecture, named Multi-Adjacency Graph Attention Network (MAGAT), we can aggregate features of near sites by considering the broad spectrum of information.

For the task at hand, we collect satellite imagery coming from the Sentinel missions~\cite{berger2012esa} and the Landsat-8 mission~\cite{roy2014landsat}. In particular, Sentinel-2A/2B and Landsat 8 satellites keep onboard multi-spectral devices (MSI) capable of acquiring 13 spectral bands for the Sentinel and 9 for the Landsat. We then pair the remotely sensed data with the on-the-ground WNV circulation dataset described in~\cite{candeloro2020predicting}, thus obtaining a dataset valid for supervised binary classification.

To prove the merits of our proposal, we conduct several experiments comparing the proposed graph aggregation with a baseline involving a single multi-band image. We also perform several ablations studies to assess the merits of different inputs, pre-training stages, and neighbourhood aggregation strategies.

\section{Related works}
\subsection{Vector-borne Disease}
In the last years, several studies applied machine learning methods in the field of vector-borne diseases and zoonoses. In~\cite{candeloro2020predicting}, the authors used derived indices derived from remotely sensed data -- \textit{i.e.}, Land Surface Temperature (LST), Normalized Difference Vegetation Index (NDVI), and Surface Soil Moisture (SSM) -- to identify the areas at risk for WNV circulation in Italy. The idea was to use data collected during the 160 days before the infection date to estimate the potential circulation of the virus two weeks in advance. To this end, the authors trained a model based on Gradient Boosting~\cite{chen2016xgboost} on epidemic data collected from 2017 to 2019.

In another line of work, the authors of~\cite{ippoliti2019defining} identified climatic and environmental eco-regions, defined as areas within which there are associations of interacting biotic and abiotic features'~\cite{bailey2004identifying}. The authors proposed to split the Italian territory into clusters (eco-climatic regions) according to seven variables, relevant to a broad set of human and animal vector-borne diseases: this way, they could highlight areas exposed to similar disease risks. Indeed, by relating the obtained results with ground-truth data, WN outbreak locations strongly end up being only a few -- four out-of-the twenty-two -- eco-regions. Such a finding is valuable in practice, as it highlights areas where surveillance measures should be prioritized. 

Concerning the applications of deep learning approaches to vector-borne diseases, most of them adapt models conceived for Computer Vision tasks. The problem in doing so lies in the large domain shift: while these backbones leverage RGB inputs, remote sensing data usually come with multiple spectral bands; therefore, the naive application of popular techniques based on Transfer Learning could lead to unsatisfactory results. In this respect,~\cite{vincenzi2021color} recommended a colourization pretext task~\cite{larsson2017colorization} to properly initialize a Deep Convolutional Neural Network (DCNN). The intuition was to exploit the high correlation between colours and the semantic characteristics of the environment (\textit{e.g.}, large bodies of water in blue, chlorophyll-rich leaves in green, and arable lands in warm tones). 

Similarly to our setting, the authors of~\cite{vincenzi2019spotting} pair satellite imagery from the Sentinel-2 mission with ground-truth information describing the presence of the vector -- in their case, \textit{Culicoles imicola}, a population of midges responsible for the transmission of the bluetongue and other orbiviruses diseases in animals -- on the ground. Differently from our work, their model focused on a single site at once but spanned across several days, thus taking into account the temporal evolution of an area in terms of climatic and environmental variables. They argued that the temporal cues can help to explain the spread of the vector: in fact, they observed that aggregating features from a recent temporal window leads to enhanced performance. 

It is noted that our work differs from~\cite{vincenzi2019spotting} in several aspects. Firstly, we target virus circulation directly instead of vector presence (\textit{i.e.}, \textit{Culicoides imicola}), which is just one of the necessary conditions influencing the viral cycle. Secondly, our analysis discards the temporal axis in favour of the spatial one: we indeed gather information from multiple nearby locations around the site of interest. However, we still maintain sensitivity to temporal cues by providing the model with \rm{the knowledge of} the month the input acquisition was carried out within.
\subsection{Graph Neural Networks (GNNs)}
In the last decade, Computer Vision tasks have greatly profited from the rise of CNNs~\cite{krizhevsky2012imagenet,he2016deep}; however, these architectures cannot handle data structured in a non-regular and complex manner, which can be naturally described as graphs. For this reason, the generalization of standard CNNs has recently gained interest, leading to the introduction~\cite{bruna2013spectral} of Graph Neural Networks (GNNs). Several works recast convolution as a message-passing operation between neighbouring nodes, which propagates features along the edges and then aggregates them to form the new representation for the pivot node. Remarkably, several efforts have also been spent on the design of tailored pooling layers~\cite{porrello2019classifying,mesquita2020rethinking}, which can be helpful for graph classification.

The authors of~\cite{defferrard2016convolutional} defined graph convolution in the spectral domain by applying Chebyshev polynomials to the graph Laplacian. On top of that work,~\cite{kipf2017semi} introduces Graph Convolutional Network (GCN), which approximates the polynomials with a truncated first-order expansion, computed on a re-normalized adjacency matrix. In~\cite{velivckovic2017graph}, the authors proposed a more sophisticated approach called Graph Attention Network (GAT), which recovers the attention mechanism to weigh the contribution of neighbouring nodes. In their proposal, the topological information is injected by masking the attention coefficients based on a binary adjacency matrix.

Unfortunately, GAT can only handle a single binary adjacency matrix: as edges serve to mask the attention coefficients properly, GAT cannot profit from real-valued edges (such as similarities between nodes). GCN~\cite{kipf2017semi} can instead do it, but its application regards only graphs with a single adjacency matrix. As discussed in the next sections, our approach addresses these limitations; in a similar vein, the authors of~\cite{gong2019exploiting} \textit{i)} exploit graphs with multi-dimensional and real-valued edges; \textit{ii)} update edges one layer after the other. To avoid exploding or vanishing values, the edge features are normalized by means of doubly stochastic normalization. Notably, their experiments on citation networks and molecular datasets show that the use of multi-dimensional edge features consistently outperforms state-of-the-art competitors. Unlike their proposal, we incorporate an initial feature extraction phase, carried by a convolutional backbone. Our rationale for doing so is to simplify the graph-based processing by utilizing high-level features instead of raw high-resolution multi-band images. Additionally, we improve the learning process by including skip connections between each graph fusion layer.

Another line of work proposes to extend the message-passing operation of GCN and GAT to multiple edges by considering inter-layer dependencies in a multi-graph structure.~\cite{shanthamallu2020gramme} proposes GrAMME, in which a specialized \textit{fusion head} combines the representation produced by different GAT heads (one for each layer of a multi-graph structure) are aggregated by specialized fusion heads. Similarly, in~\cite{zangari2021graph} the authors propose ML-GCN and ML-GAT to extend the graph convolution with inter-graph connections. These solutions differ from our proposal, as they cannot take advantage of both GAT- and GCN-based processing simultaneously. Moreover, the aggregation mechanism of~\cite{zangari2021graph} is designed for multi-graph structures in which both a separate adjacency matrix and a set of node features for each layer are defined; instead, our case study provides only a single set of node features. Nevertheless, for a fair comparison against our proposal, in Sec.~\ref{sec:other_gnns} we introduce a multi-layer GNN baseline inspired by GrAMME-Fusion~\cite{shanthamallu2020gramme} and ML-GCN~\cite{zangari2021graph}.

\medskip
\noindent\textbf{Application of GNNs to remote sensing data.}~In the EO field, GNNs have so far been employed in few works. The authors of~\cite{wu2021multiscale} propose a change detection model combining a multiscale segmentation technique and a GCN: the former extracts object-wise high-level features from the multiscale input images; then, for each scale, the features are arranged as the nodes of a graph, with the edges representing the spatial relationships between nearby objects. The graphs are then processed by GCNs and merged into a single one by a fusion model, producing the change detection result.

For the task of land cover classification, the authors of~\cite{censi2021attentive} propose an attentive spatial-temporal GNN (STEGON) to model both the temporal and spatial aspects that characterize satellite data. Their method involves an initial 1-d CNN to extract relevant temporal patterns, followed by a graph aggregation technique via GAT and self-attention to incorporate the information about the spatial neighbourhood of the target node. With their work, the authors emphasize the significance of considering both time and space dimensions when dealing with satellite data.

In~\cite{ouyang2021combining}, the authors point out that common semantic segmentation approaches overlook the strong correlation between different classes: for example, bridges tend to be closer to rivers and far from cultivated fields. On this basis, they propose to model such relations through a dedicated graph, whose nodes are the clusters provided by the superpixel segmentation algorithm~\cite{li2015superpixel}; edge similarities are instead formulated in terms of average colour similarity in input space. Finally, a GCN module performs the classification of each superpixel.

\section{Problem setting}
We aim to identify the conditions favouring WN virus circulation, which could enable the implementation of targeted surveillance plans. As mentioned, various works~\cite{tran2014environmental,conte2015spatio} have already proven the connection between the West Nile Virus circulation and several environmental factors, such as vegetation, temperatures, and water coverage. We build upon that relationship and exploit herein a large amount of satellite data produced in the last years: differently from previous studies based on classical statistical tools~\cite{candeloro2020predicting}, we investigate whether the presence/absence of the WNV circulation in Italy can be inferred through deep learning-based approaches.

In the following, we describe the dataset we have forged to pursue our aim: basically, it pairs data describing the WNV circulation on the ground (see Sec.~\ref{sec:westnile}) with satellite imagery (see Sec.~\ref{sec:imagery}).
\subsection{Data on West Nile Virus circulation}
\label{sec:westnile}
\begin{figure}
    \centering
    \includegraphics[width=0.9\columnwidth]{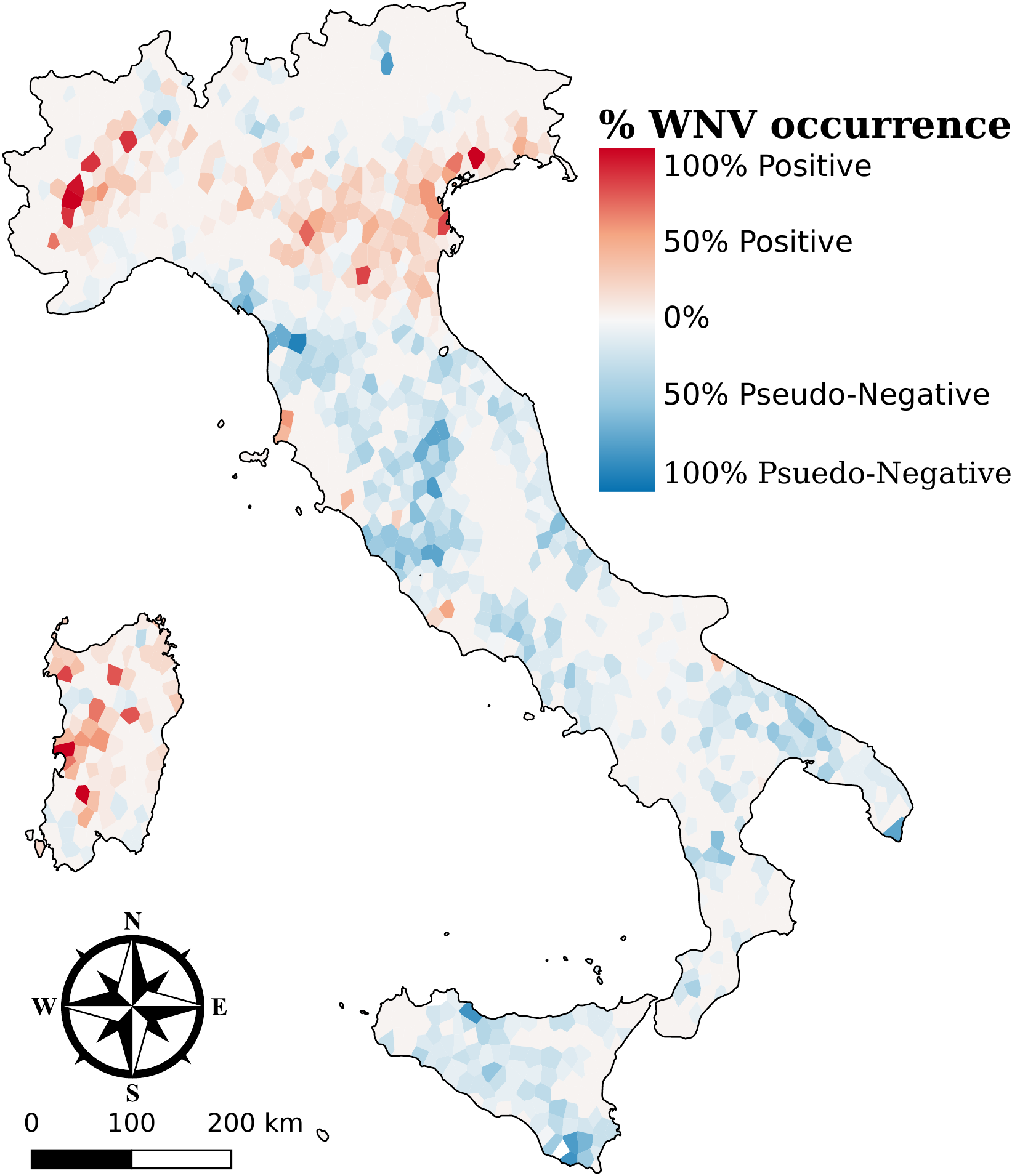}
    \caption{\textbf{Distribution of ground-truth data} on the Italian peninsula. For better visualization, the individual occurrences are spatially clustered. We mark the prevalence of WNV occurrence (positive samples - red) over pseudo-absence (negative samples - blue). As can be seen, while the negative samples are spread across the territory, positive cases are more concentrated, indicating a strong correlation with the local environment (\textit{best seen in color}).}
    \label{fig:posneg}
\end{figure}
We first gathered the veterinary dataset described in~\cite{candeloro2020predicting}, collecting observations about the virus circulation in Italy from 2008 to 2019. Briefly, the authors collected ground truth information regarding the geographical locations (coordinates) of the virus outbreaks, detected by the National Disease Notification System of the Italian Ministry of Health\footnote{SIMAN, \url{www.vetinfo.it}}~\cite{colangeli2011sistema}. Specifically, we focus our analysis on the years 2017, 2018, and 2019 and consider \textit{positive sites} all those places where the presence of the virus was observed in either equids, birds, or mosquitoes. During those years, data have been recorded using the same methodology; any possible source of inconsistency (geographical location, sampling date, etc.) was verified and corrected.

This way, we could gather information only about confirmed positive instances. Since the surveillance activities were not performed in all Italian territories (but only in those areas with historical evidence of virus circulation or at higher risk of WNV introduction), the indication of which places can be considered as negative sites is not always available~\cite{candeloro2020predicting}. For this reason, pseudo-absence data were randomly generated and distributed in areas suitable for the virus presence -- such as places characterized by an altitude under 600 meters and located in medium inhabited centers -- where the disease was never reported in the past.

\begin{figure}
    \centering
    \includegraphics[width=0.95\linewidth]{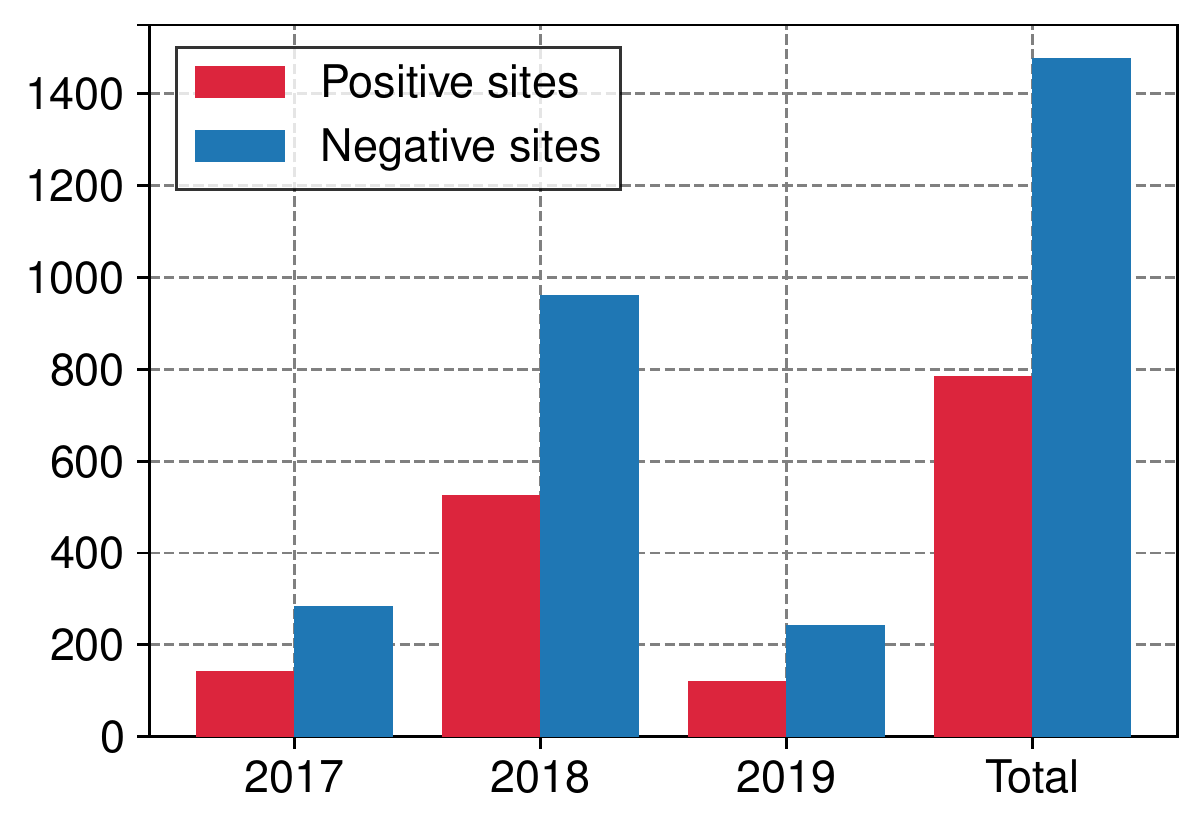}
    \caption{\textbf{Number of recorded cases} -- positive occurrences and pseudo-absence data -- for each year.}
    \label{fig:dataset}
\end{figure}

To sum up, the dataset we collect comprises 2264 cases, split into 786 positives and 1478 negatives. To provide a visual description, we show the spatial distribution of positive and \textit{pseudo-negative} sites in Fig.~\ref{fig:posneg}. Instead, we refer the reader to Fig.~\ref{fig:dataset} for the temporal data distribution over the years.
\subsection{Satellite Imagery}
\label{sec:imagery}
\begin{table}
\caption{Spectral bands shared by Sentinel-2 and Landsat-8, with corresponding wavelength ($\mu$\MakeLowercase{m}) and spatial resolution (\MakeLowercase{m}).}
\label{tab:sat_bands}
    \begin{center}
        \begin{tabular}{l l c c}
        \toprule
        & \mytabheader{Bands} & \mytabheader{Wavelength} & \mytabheader{Res (Sentinel/Landsat)}\\
        \midrule
        $\text{B}_{1}$ & Coastal Aerosol & 0.443 & 60/30 \\
        $\text{B}_{2,3,4}$ & BGR channels & 0.490 & 10/30 \\
        $\text{B}_{8\text{A}}$ & Vegetation Red Edge & 0.865 & 20/30 \\
        $\text{B}_{11}$ & SWIR & 1.610 & 20/30 \\
        $\text{B}_{12}$ & SWIR & 2.190 & 20/30 \\
        \bottomrule
        \end{tabular}
    \end{center}
\end{table}
In the following paragraphs, we describe the main sources of remote sensing data, which we pair with the aforementioned information describing WNV circulation. These data are either in the form of raw spectral bands~\cite{calzolari2015west,paz2015climate} as well as specific measurements (\textit{e.g.}, temperature and soil moisture) already acknowledged to partially explain WNV circulation.

Firstly, we aim to represent each geographical site with a multi-band satellite image depicting that site at a certain time. In this respect, we get the data captured by both Sentinel-2 and Landsat-8 satellites:

\noindent\textbf{Sentinel-2.}~The two twin satellites Sentinel-2A and 2B, launched in 2015 and 2017 respectively in the frame of the European program Copernicus, acquire 13 spectral bands at different spatial resolutions (10, 20, and 60 meters per pixel), with a revisit time equals to five days.

\noindent\textbf{Landsat-8.}~Landsat-8 satellite, launched in 2013, carries onboard the Thermal Infrared Sensor (TIRS) and the Operational Land Imager (OLI), acquiring respectively two and nine spectral bands. The satellite flies in a near-polar orbit, at an altitude of 705 Km, and acquires images of the same territory every 16 days. For this work, we consider the nine spectral bands captured by the OLI instruments, having a spatial resolution of 30 meters per pixel.

\medskip
We limit our analysis to the seven bands these two systems have in common (see Tab.~\ref{tab:sat_bands}). As in~\cite{vincenzi2019spotting}, we adopt 20 meters as the default spatial resolution and resize the available tiles to account for the different resolutions. We collect a total of 37876 satellite images, which constitute the basis for both train and test sets. Each of these is paired with the temporally closest ground-truth observation; therefore, the same observation can be associated to multiple satellite images.

\noindent\textbf{Land Surface Temperature (LST).}~It represents a fundamental determinant of the terrestrial thermal behaviour, as it controls the effective radiating temperature of the Earth's surface. LST was captured separately during day and night from a dual-view scanning temperature radiometer, an instrument onboard the Copernicus Sentinel-3A and 3B satellites~\cite{fletcher2012sentinel}. They fly in low Earth orbit at 800-830 km of altitude, feature a daily revisit time over the same place, and have a spatial resolution of one km.

\noindent\textbf{Surface Soil Moisture (SSM).}~It represents the percentage of the relative water content of the top few centimeters of the soil, indicating how dry or wet it is. Moreover, it also provides insights into local precipitation impacts and soil conditions. The SSM data have been retrieved from the Copernicus Global Land Service~\cite{bauer2018toward}, which includes two polar-orbiting satellites that operate day and night performing C-band synthetic aperture radar imaging; this ensures the acquisition regardless of the weather. Similar to Sentinel-3, Sentinel-1 has a daily revisit time and a spatial resolution of one km.

\begin{figure*}
    \centering
    \includegraphics[width=0.95\textwidth]{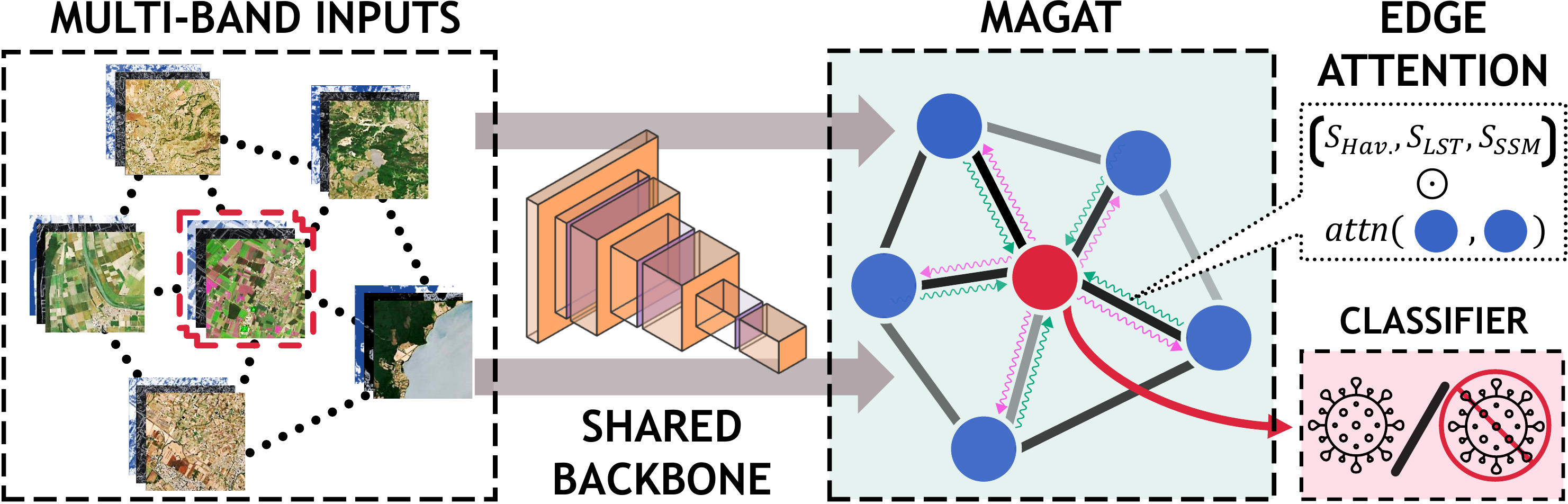}
    \caption{An \textbf{overview of the proposed pipeline}. 1) A target multi-band sample and its neighbours are independently passed through a shared convolutional network; 2) The obtained high-level features define the content of each node of a graph; the edges, instead, encode several distances; 3) The information is propagated by means of our \textbf{Multi-Adjacency Graph Aggregation Network (MAGAT)}, which builds upon an attention function over the nearby nodes; 4) The final decision is obtained by a simple linear projection of the target's node features.}
    \label{fig:framework}
\end{figure*}
\section{Model}
\label{sec:overview}
\noindent\textbf{Overview.}~We look for a way to exploit neighbour sites while analysing a given infection point. To this aim, we propose a model composed of three main blocks: a features extractor (Sec.~\ref{sec:feaext}), a graph aggregation module, and a final classifier (Sec.~\ref{sec:graph_agg}). An overview of the architecture is shown in Fig.~\ref{fig:framework}.
\subsection{Features Extractor}
\label{sec:feaext}
The role of the feature extractor is to process the multi-band and high-resolution satellite images to obtain a lower dimensional feature vector. During this step, redundant or noisy information is discarded, leaving only high-level details that better summarize the input patch.

\medskip
\noindent\textbf{Architecture.}~We adopt ResNet18~\cite{he2016deep} as backbone, which comprises four blocks with two residual units each. The presence of residual connections prevents the representations from degrading towards the end of the network; moreover, as clarified in~\cite{kolesnikov2019revisiting}, ResNet-based networks are more suitable for self-supervised representation learning.

\medskip
\noindent\textbf{Pre-training.}~When dealing with a constrained number of images, common approaches involve Transfer Learning techniques to boost the quality of the representations extracted by the model. Among these, the leading approach usually involves an initial pre-train stage on a large dataset, followed by fine-tuning (knowledge transfer) on the target data~\cite{he2017mask,ren2015faster}. In the former phase, the dataset commonly adopted is ImageNet~\cite{deng2009imagenet}, a large image classification dataset. However, since it only features RGB images, its application appears less appropriate for the task at hand. Indeed, its use discards potentially meaningful information present in multi-spectral satellite imagery. 

Instead, we rely on a tailored solution to exploit the additional information provided by satellite data. Following~\cite{vincenzi2021color}, we construct a colourization pretext task: we require the model to reconstruct the RGB spectral bands while taking as input the other ones. Starting from BigEarthNet~\cite{sumbul2019bigearthnet,sumbul2021bigearthnet} -- an open-access dataset containing 590000 labelled (with land cover type) Sentinel-2 images -- we train a ResNet-based autoencoder architecture on the colourization task. Once the model has reached good capabilities, we discard the decoder and keep the encoder parameters, which are used to initialize the feature extractor that will be trained to solve the WNV classification task.

\medskip
\noindent\textbf{Conditional Batch Normalization.}~As discussed in ~\cite{candeloro2020predicting}, the yearly distribution of WNV infections exhibits a clear seasonal trend, highly affecting the distribution of positive occurrences. To capture such a pattern, we include the information of the relative timestamp as an additional covariate of our study. Specifically, we modulate the convolutional feature maps based on the month in which the input image was captured by satellites. In practice, we replace each Batch Normalization (BN) layer of the feature extractor with its counterpart Conditional Batch Normalization (CBN)~\cite{courville2017modulating}. 

In practice: while classical BN reckons on a single set of learnable affine parameters, the CBN layer we use provides 12 sets of parameters, one for each month. Therefore, given a batch $\mathcal{B} = \{ \matrixsym{H}^{i}\}_{i=1}^{N}$ of $N$ samples, where $\matrixsym{H}\in\mathbb{R}^{C\times W\times H}$ denotes a generic feature map produced from a convolutional layer, CBN on channel $c\in\{1,\ldots,C\}$ and location $(w, h)\in\{1,\ldots,W\}\times\{1,\ldots,H\}$ is defined as:
\begin{equation}
    \operatorname{CBN_c}(H^i_{c,h,w} | \gamma_c, \beta_c) = \gamma_c\frac{H^i_{c, w, h} - \mathbb{E}_{\mathcal{B}}[\matrixsym{H}_{c, \cdot, \cdot}]}{\sqrt{\operatornamewithlimits{Var}_{\mathcal{B}}[\matrixsym{H}_{c, \cdot, \cdot}]+\epsilon}} + \beta_c,
\end{equation}
where $\epsilon$ is a constant that prevents numerical issues and $\gamma_c, \beta_c$ are learnable parameters controlling the affine transformation the normalized feature maps are subject to.
\subsection{Graph Aggregation}
\label{sec:graph_agg}
As previously mentioned, the density and spread of the vectors that carry WNV are closely related to climatic and environmental characteristics. To model these relations, we arrange the features produced by neighbouring sites as nodes in a graph and apply a Graph Neural Network (GNN).

\medskip
\noindent\textbf{Overview.}~Let $N$ be the number of nodes in the graph, each with $F^0$-dimensional features. We then indicate with $\matrixsym{X}^0\in\mathbb{R}^{N\times F^0}$ the input node features and with $\matrixsym{S}^0\in\mathbb{R}^{N\times N\times G}$ its multi-valued affinity matrix ($G$ stands for the number of different input relations). In our experiments:
\begin{itemize}
    \item \textit{Nodes}. We forge each input graph by sampling a random location -- called \textbf{pivot} node -- and then gathering the 10 spatially closest neighbours locations. 
    \item \textit{Edges}. We provide a stack of three matrices: the geographical (Haversine) distance and the two other ones based on LST and SSM (namely,for each band we compute the mean absolute pair-wise distance between the respective the nodes, thus outlining the average difference in temperature and soil moisture between the locations). All those distances are converted to similarities (through a Gaussian kernel with $\sigma=1$) and then further pre-processed via double-stochastic normalization (discussed in the next paragraphs).
\end{itemize}

The building block of our proposed architecture consists of a multi-head residual graph layer -- termed MAGAT -- which basically extends GAT~\cite{velivckovic2017graph} for leveraging multiple adjacency matrices. Briefly, each head operates on a separate adjacency matrix $\matrixsym{S}^0_{\cdot,\cdot,g}$ to process the input features $\matrixsym{X}^0$. Results are then aggregated to produce two results: a new $N\times F^1$ nodes' representation $\matrixsym{X}^1$, with $F^1\triangleq G\cdot F^{Int}$, and $N\times N\times G$ edge feature matrices $\matrixsym{S}^1$. We stack multiple of these layers, each of which is fed with the affinity matrices produced by the previous layer.

After having applied these transformations, we discard the neighbours and focus only on the pivot node; this way a single vector embedding is fed to the final classification layer. The latter outputs the prediction for the presence of WNV circulation.

\medskip
\noindent\textbf{MAGAT layer.}~In formal terms, let us first review the similarity between nodes $i$ and $j$ as originally conceived by the authors of GAT:
\begin{equation}
  \label{eq:attn_function}
  s(\matrixsym{X}^0_i,\matrixsym{X}^0_j) = \exp\left\{\mathrm{LReLU}\left(\matrixsym{p}^T[\matrixsym{V}X^0_{i}\Vert \matrixsym{V}X^0_{j}]\right)\right\},
\end{equation}
where $V\in\mathbb{R}^{F^{Int}\times F^0}$ linearly projects the input features to an intermediate dimension $F^{Int}$, $\Vert$ stands for the concatenation operator, and $p\in\mathbb{R}^{2F^{Int}}$ is a learnable vector that allows for a single scalar output. In this simple formulation, the only way to inject an external adjacency matrix is by zeroing out the contribution from unconnected nodes. Drawing inspiration from~\cite{gong2019exploiting}, MAGAT learns a separated similarity on top of each matrix $\matrixsym{S}^0_{\cdot,\cdot,g}$ provided as input. Formally, we modify Eq.~\ref{eq:attn_function} by introducing an attenuation factor that emphasizes the contributions from similar nodes in the neighbourhood:
\begin{equation}
  \label{eq:alpha-2}
  \hat{\matrixsym{\alpha}}_{i,j,g} = S_{i,j,g}^0 \cdot s(\matrixsym{X}^0_i,\matrixsym{X}^0_j),
\end{equation}
where:
\begin{equation}
    \matrixsym{S}^0 = \begin{cases}
    \matrixsym{S}^0_{i,j}, & \text{if $j\in Neighbours(i)$}\\
    0, & \text{otherwise}\end{cases}.
\end{equation}

In our approach, the resulting attention coefficients become the input affinity matrices for the subsequent layer. However, the repeated application of Eq.~\ref{eq:alpha-2} may result in severe numerical instabilities. To avoid such an issue, we apply double-stochastic normalization on $\hat{\matrixsym{\alpha}}$ by means of the Sinkhorn–Knopp iterative algorithm~\cite{sinkhorn1967concerning,adams2011ranking}. This way, we obtain the final normalized affinity matrices: for each $g=1,2,\ldots,G$ we have a square non-negative real matrix with columns and rows summing to 1. 

Once the final attention coefficients $\matrixsym{\alpha}$ have been computed, we perform the node feature aggregation on each affinity matrix. Specifically, we define a multi-head structure in which the $G$ heads perform a separate message-passing operation. This strategy ensures an efficient feature aggregation since each head can compute its output in parallel. In our design, each $g=1,2,\ldots,G$ sub-layer computes an initial linear combination of the node features $\matrixsym{X}^0$, then scales the result for $\matrixsym{\alpha}_{\cdot,\cdot,g}$. The independent results are then concatenated to obtain $\matrixsym{X}^1$. Formally, we compute:
\begin{equation}
  \label{eq:attention}
  \hat{\matrixsym{X}} =\bigparallel_{g=1}^G\left(\matrixsym{\alpha}_{\cdot,\cdot,g}\matrixsym{W}_g\matrixsym{X}^0\right),
\end{equation}
where $\matrixsym{W}_g\in\mathbb{R}^{F^{Int}\times F^0}$ is a learnt parameter matrix and $\bigparallel$ defines the concatenation operation.

Finally, we provide a residual connection between the input node features and the output of MAGAT, whose purpose is to prevent vanishing gradients thus easing the training of the model. In practice, if indicating with $u:\mathbb{R}^{F^0}\rightarrow\mathbb{R}^{F^1}$ a single linear projection layer, the MAGAT layer then computes:
\begin{equation}
  \label{eq:residual}
  \matrixsym{X}^1 = \sigma(\hat{\matrixsym{X}}) + u(\matrixsym{X}^0),
\end{equation}
where $\sigma$ is the ELU~\cite{clevert2015fast} activation function. 
\subsection{Model ensemble}
\label{sec:ensemble}
Following~\cite{vincenzi2021color}, we further take advantage of both spectral bands (\textit{i.e.}, those being outside the visible spectrum) and the RGB ones. In particular, we set up an ensemble model that averages the predictions from two models: one is fed with RGB inputs and is pre-trained on the ImageNet dataset; the other takes spectral bands as input and leverages colourization pre-training as was proposed in~\cite{vincenzi2021color}.

\section{Experiments}
\begin{table}
\caption{Comparison between our graph-based approach and a baseline featuring only the information available for the target node. The superior performance of MAGAT indicates the advantages of additional information from surrounding environment.}
\label{tab:res_v_magat}
\begin{center}
    \begin{tabular}{ccccccc} 
        \toprule
        & \textbf{Domain} & \textbf{Pre-train.} & \textbf{Acc.} & \textbf{Pr.} & \textbf{Rc.} & \textbf{F1} \\
        \midrule
        \multirow{3}{*}{ResNet18} & RGB & ImageNet & .825& .722& .858& .780\\
        & Spectral & Colorization & .840& .757& .861& .795\\
        & Both & Ensemble & \underline{.855} & \underline{.761} & \underline{873}  & \underline{.813} \\
        \midrule
        \multirow{3}{*}{MAGAT} & RGB   & ImageNet    & .921& .834& .973& .897\\
        & Spectral & Color.\ & .922& \textbf{.850}    & .952& .898\\
        & Both & Ensemble\ & \textbf{.926}    & .844& \textbf{.977} & \textbf{.905}    \\
        \bottomrule
    \end{tabular}
\end{center}
\end{table}
In this section, we present the results obtained in the scenario presented in Sec.~\ref{sec:westnile} about West Nile Virus circulation. We first summarize the choices made to set up the experimental setting; then, several comparisons are proposed to highlight the contribution of each introduced component.
\subsection{Experimental details}
\medskip
\noindent\textbf{Pre-processing.}~For each location included in the dataset, we sample a squared satellite patch centered around its spatial coordinates (latitude and longitude). We interpolate each band to obtain a resolution of 20 meters, thus aligning different bands and obtaining a homogeneous format. All those pixels marked as either invalid, saturated, or related to heavy clouds are set to a pre-defined ``NoData'' value: in this respect, we discard the examples presenting a ratio of NoData pixels higher than an acceptance threshold (set to 10\% in our experiments). 

\medskip
\noindent\textbf{Hyperparameters.}~We set the batch size to 16 and optimize for 20 epochs with plain Stochastic Gradient Descent (SGD). To reduce the impact of overfitting, we use dropout before the final classification layer (with drop probability equal to $0.2$). We minimize the binary cross-entropy loss during training.

\medskip
\noindent\textbf{Validation.}~Due to the class unbalance present in our dataset ($65.3\%$ of the samples are associated with a negative label), we resort not only to accuracy but also to proper metrics, such as \textit{precision}, \textit{recall}, and \textit{F1-score}.

We split the dataset into two disjoint sets for training and inference: the former comprises data from 2017 and 2018, while the test phase performs on observations from 2019. As the split is made on the time dimension, a certain geographical location may appear in both train and test sets due to the periodical nature of satellites' motion. Finally, all the results have been obtained by repeating each experiment five times and then reporting the average. 
\subsection{The value of neighbouring nodes}
We start the experimental investigation of our graph-expanded approach by comparing it with a very simple baseline, which does not involve graph learning at all and considers one location at once: this way, we can demonstrate the benefits of information from neighbouring places. The baseline at stake is ResNet18, which we initialize according to the same pre-train strategies discussed in Sec.~\ref{sec:overview}. We refer the reader to Tab.~\ref{tab:res_v_magat}, which reports the results of such a comparison. As can be observed, we report the performance for different input modalities\footnote{For the rest of the article the term "\textit{Spectral}" is meant to indicate the subset of the bands not including the visible part of the spectrum}.

We can draw the following conclusions: firstly, the proposed approach is better at estimating the presence/absence of the WNV circulation, reaching a top accuracy of $92.6\%$ and an F1-score of $90.5\%$ for the ensemble strategy. Such evidence confirms our intuition, according to which the further consideration of the surrounding environment -- embodied by neighbouring nodes in our setting -- could prove beneficial for shaping the trend of WNV circulation. 

Secondly, while exploiting the spectral domain provides a consistent gain in terms of performance for the baseline approach, we notice that this does not hold for MAGAT. In fact, when leveraging graph-learning, the results deriving from the two domains end up comparable. Namely, the extra information provided by the parts of the spectrum outside the visible appears less critical when considering multiple adjacent nodes. We conjecture that the message-passing operation could indeed mitigate the shortcomings of RGB, introducing novel sources of information.

Thirdly, our results confirm what was said in~\cite{vincenzi2021color}: as ensembling the two domains provides the best performance for both approaches, the involved pre-training strategies seem to deliver different but valuable concepts.
\subsection{Benefits of deep approaches}
\begin{table}
\caption{Comparison between shallow classifiers and our deep proposal MAGAT.}
\label{tab:ml_v_magat}
    \begin{center}
        \begin{tabular}{l c c c c c } 
        \toprule
                                        & \textbf{Pre-train.} & \textbf{Acc.} & \textbf{Pr.} & \textbf{Rc.} & \textbf{F1}\\
        \midrule
            Logistic Regression               & \xmark            & .717             & .571              & .869             & .689             \\
            Random Forest                     & \xmark            & .811             & .675              & .917             & .778             \\
            Gradient Boosting                 & \xmark            & \underline{.837} & \underline{.710}  & \underline{.926} & \underline{.804} \\
        \midrule
        \multirow{3}{*}{MAGAT}             & ImageNet          & .921             & .834             & .973             & .897             \\
                                            & Color.            & .922             & \textbf{.850}    & .952             & .898             \\
                                            & Ensemble & \textbf{.926}    & .844             & \textbf{.977}    & \textbf{.905}    \\
        \bottomrule
        \end{tabular}
    \end{center}
\end{table}
As deeply discussed, we rely on deep learning techniques and multi-band satellite imagery to address the task at hand. However, the actual benefits of such a demanding approach could be argued: could simpler solutions perform comparable or even better but at lower costs?

We herein attempt to answer these doubts by comparing MAGAT against several shallow classifiers that leverage hand-crafted features, such as:
\begin{itemize}
    \item \textbf{Logistic regression}~\cite{cox1958regression} learns a linear decision boundary. It does so by maximizing -- via gradient descent -- the log-likelihood on the training set.
    \item \textbf{Random forest}~\cite{breiman2001random} is a \textit{bagging} ensemble algorithm: a set of simple models are trained independently on subsets of the original data. During inference, the predictions of these models are combined via majority voting.
    \item \textbf{Gradient boosting}~\cite{friedman2001greedy} is based on \textit{boosting}, an ensemble technique that trains a cascade of models, each of which focuses mainly on the examples misclassified by previous ones. The overall prediction is a weighted sum of the outputs of each model.
\end{itemize}
These methods cannot be fed with high-resolution multi-band images; therefore, we provide them with the average values of the spectral bands introduced in Sec.~\ref{sec:imagery}, along with the geographic coordinates and the NDVI. The latter is computed from the $\text{B}_{8\text{A}}$ and $\text{B}_{4}$ bands (see Tab.~\ref{tab:sat_bands} for reference).

We report the results for such a comparison in Tab.~\ref{tab:ml_v_magat}. Among the shallow approaches, gradient boosting leads to higher performance. However, our approach leads by a margin of $10\%$ on the F1 score. Such a gap is also clear when looking at the precision and recall metrics: all the shallow classifiers tend to overemphasize the positive class, thus presenting a high recall but at the cost of low precision.

While we focus on finding the best performer for the task, we still highlight that classical machine learning methods feature faster inference and less expensive training compared to their deep learning counterparts, which could favor them in scenarios with stricter constraints.
\subsection{Impact of Conditional Batch Normalization}
\label{subsec:cbn}
\begin{figure}
    \includegraphics[width=0.9\columnwidth]{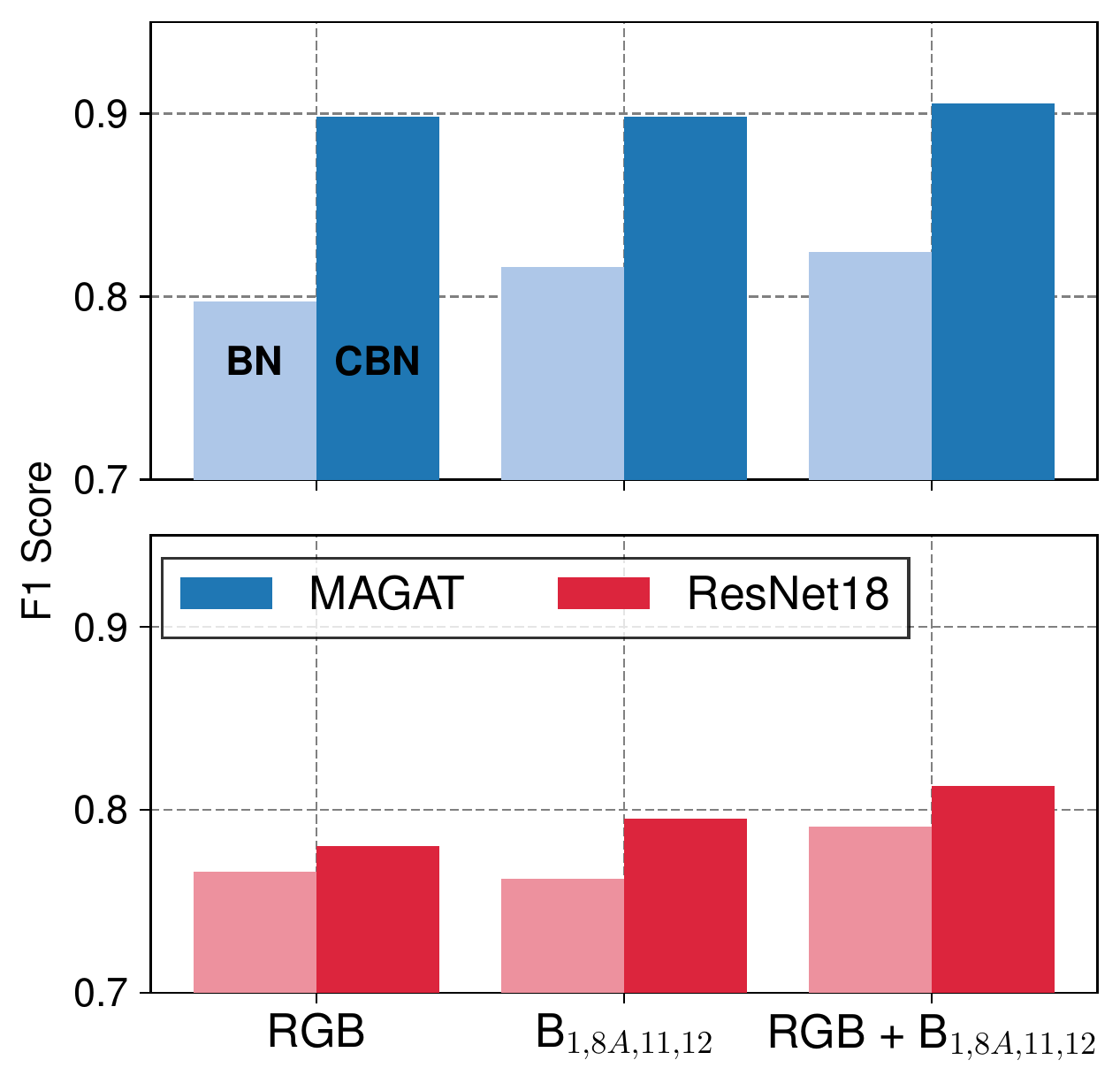}
    \caption{Barplot showing the \textbf{impact} in terms of F1 Score \textbf{of Conditional Batch Normalization} (CBN) over its naive counterpart (BN).}
    \label{fig:bn_vs_cbn}
\end{figure}

As discussed in~\cite{conte2015spatio}, in addition to being affected by environmental factors such as temperature, humidity, and soil moisture, the WNV circulation is also governed by seasonal patterns. In light of this, we introduced the Conditional Batch Normalization layer to provide the acquisition month to the feature extractor. Fig.~\ref{fig:bn_vs_cbn} shows that such a design choice leads to a higher F1 score; moreover, the gain is consistent across different choices (\textit{i.e.}, backbone and input modality).
\subsection{Impact of different similarities in MAGAT}
\label{subsec:impact_similarity}
\begin{table}
\caption{Impact of different edge cues on performance.}
\label{tab:adjacency_comparison}
    \begin{center}
        \begin{tabular}{lccccc} 
        \toprule
        \textbf{Inputs}               & \textbf{adjacency}    & \textbf{Acc.} & \textbf{Pr.}  & \textbf{Rc.}  & \textbf{F1} \\
        \midrule
        \multirow{5}{*}{RGB}          & \textit{uniform}         & .904          & \textbf{.855} & .884          & .869 \\
                                        & LST                   & .890          & .818          & .894          & .854 \\
                                        & SSM                   & .900          & .846          & .885          & .865 \\
                                        & LST + SSM             & .889          & .825          & .879          & .851 \\
                                        & \textit{all} (ours)   & \textbf{.920} & .834          & \textbf{.973} & \textbf{.897} \\
        \midrule
        \multirow{5}{*}{Spectral}       & \textit{uniform}         & .889          & .788          & .948          & .861 \\
                                        & LST                   & .880          & .775          & .942          & .850 \\
                                        & SSM                   & .887          & .786          & .946          & .859 \\
                                        & LST + SSM             & .901          & .805          & \textbf{.960} & .876 \\
                                        & \textit{all} (ours)   & \textbf{.922} & \textbf{.861} & .952          & \textbf{.898} \\
        \midrule
        \multirow{5}{*}{Both} & \textit{uniform}         & .909          & .830          & .942          & .883 \\
                                        & LST                   & .905          & .830          & .928          & .876 \\
                                        & SSM                   & .904          & .806          & .968          & .879 \\
                                        & LST + SSM             & .907          & .824          & .944          & .880 \\
                                        & \textit{all} (ours)   & \textbf{.926} & \textbf{.844} & \textbf{.977} & \textbf{.905}  \\
        \bottomrule
        \end{tabular}
    \end{center}
\end{table}
Here, we empirically review alternative strategies to build the input affinity matrices $\matrixsym{S}$. In particular, we compare ours (termed \textit{all}, which comprises geographic, LST, and SSM distances) with using either only LST and SSM similarities. Additionally, we also wonder what happens when providing no informative edges at all: in this respect, we set up a baseline called \textit{uniform} that assigns the same importance to every edge (it ends up using node features solely, as hold in Eq.~\ref{eq:alpha-2}).

Tab.~\ref{tab:adjacency_comparison} reports the results of this investigation for different modalities. Overall, using all the available information as advocated by our proposal consistently yields higher F1 scores. However, we also note that the \textit{uniform} adjacency leads to competitive results: in 2 out of 3 cases, it even surpasses the combination of environmental bands (LST + SSM). 

Furthermore, considering only the surface temperature or soil moisture seems misleading for the model. In fact, without providing any hints about the geographical distance between nodes, locations sharing similar environmental features may be suspected of virus transmission even if they are too far apart for the virus to spread.
\subsection{The relevance of pre-training}
\begin{table}
\caption{Impact of different pre-training \textit{vs} training from scratch.}
\label{tab:pret_ablat}
    \begin{center}
        \begin{tabular}{lccc} 
        \toprule
        \textbf{Inputs} & \textbf{Pre-training} & \textbf{ResNet18} & \textbf{MAGAT}\\
        \midrule
        \multirow{2}{*}{RGB}          & \xmark             & .704  & .740  \\
                                        & ImageNet           & .780  & .897  \\
        \midrule                     
        \multirow{2}{*}{Spectral}       & \xmark             & .716  & .867  \\
                                        & Color.\            & .795  & .898  \\
        \midrule
        \multirow{2}{*}{Both} & \xmark             & .747  & .808  \\
                                        & ImageNet + Color.\ & .813  & .905  \\
        \bottomrule
        \end{tabular}
    \end{center}
\end{table}
Due to the limited size of the dataset employed, our solution exploits different pre-train strategies (\textit{e.g.}, based on the ImageNet dataset~\cite{olga2015imagenet}) to mitigate overfitting. We still question the relevance of that detail, hence evaluating the results of MAGAT with a randomly initialized feature extractor. Unsurprisingly, Tab.~\ref{tab:pret_ablat} highlights a clear advantage in using a pre-trained network: when evaluating both Colourization and RGB pre-train (last row) the ResNet18 and MAGAT ensemble models show an improvement of $6.6$ and $9.7$ respectively. 

We also note that, when pre-training is not involved, the performance of the MAGAT ensemble is inferior to the average performance of the models trained on the RGB and \bspec~bands separately. Such a result indicates that -- without a proper initialization strategy -- there are no advantages in combining the features learned by the two models.
\subsection{Effectiveness of MAGAT against GNN baselines}
\label{sec:other_gnns}
\begin{figure*}
    \centering
    \includegraphics[width=0.95\textwidth]{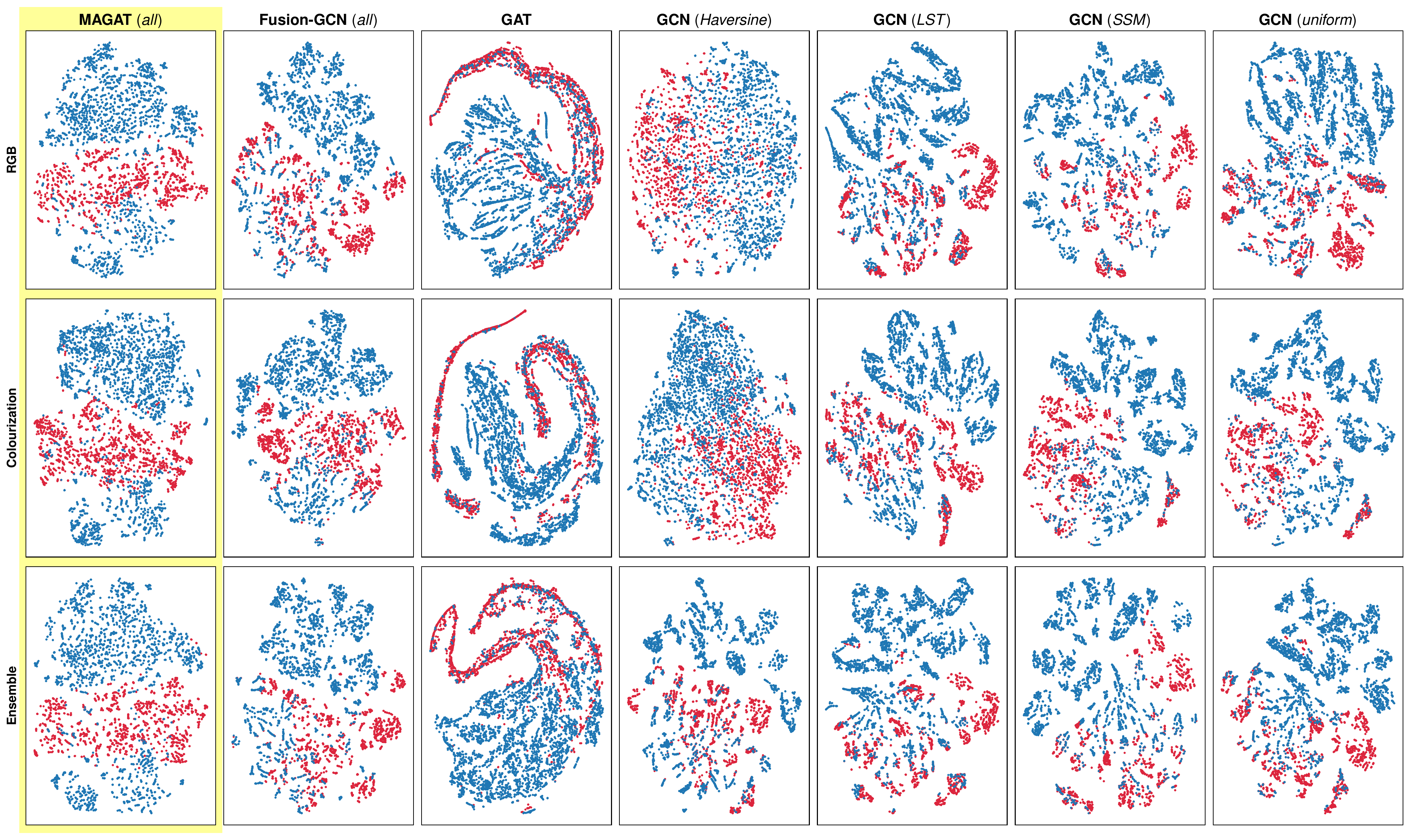}
    \caption{Depiction of the \textbf{feature spaces spanned by different GNNs} after t-SNE~\cite{vandermaaten2008visualizing} compression. As can be seen, the combined aggregation of all edge features in MAGAT results in less overlapping between the positive samples (\textcolor[HTML]{DC253D}{red}) over the pseudo-negatives (\textcolor[HTML]{1F77B4}{blue}) (best seen in color).}
    \label{fig:tsne}
\end{figure*}
Once we have discussed the importance of the various parts of MAGAT, we now wish to assess its performance under the light of other established GNN-based approaches. We consider both GAT and GCN as baselines due to their widespread use in literature. For the latter, we evaluate all choices of single-dimensional adjacency matrix (\textit{uniform}, LST, SSM, and geographic distance). Secondly, inspired by GrAMME-Fusion~\cite{shanthamallu2020gramme}, we include a multi-adjacency extension of GCN, which we call \textbf{Fusion-GCN}; herein, a \textit{fusion} layer combines the contribution of multiple independent GCN layers as follows:
\begin{equation}
    \mathbf{X} = \sum_{g=1}^{G}{\beta_g \mathbf{z}_g},
\end{equation}
with $\mathbf{z}_g$ being the output of the GCN layer related to the $g$\textsuperscript{th} adjacency matrix.

We first propose a quantitative comparison in Tab.~\ref{tab:gat_gcn_vs_magat}: as can be seen, our proposal MAGAT reaches stronger results w.r.t.\ other GNN-based approaches under all combinations of pre-train and adjacency matrix. This finding demonstrates the effectiveness of MAGAT in capturing the temporal and spatial dependencies of complex graph-structured data. Interestingly, the poor result of GAT indicates a clear advantage to using the information regarding the edges, which is in line with our previous findings of Sec.~\ref{subsec:impact_similarity}.

In addition to the quantitative analysis, in Fig.~\ref{fig:tsne} we provide a qualitative assessment of the feature spaces generated by the different GNN models. Inspired by other works~\cite{kipf2017semi,velivckovic2017graph,shanthamallu2020gramme}, we utilize t-SNE to reduce and visualize the feature spaces generated by the different models. Our observations reveal that MAGAT produces a feature representation that better separates the different classes, providing additional evidence for the effectiveness of our approach in capturing the underlying structures in the data. Additionally, the representation produced by GAT suggests a poor separation between the classes, which negatively affects the overall performance of the model as shown in the previous table. This effect is less pronounced in the other GCN-based baselines, highlighting the importance of incorporating the environmental and climatic relationships between nearby locations.
\begin{table}
\caption{Impact of MAGAT against other GNNs, measured as F1 score.}
\label{tab:gat_gcn_vs_magat}
\begin{center}
    \begin{tabular}{lcccc}
    \toprule
    \textbf{Inputs}               & \textbf{adjacency}    & \textbf{RGB}  & \textbf{Spectral} & \textbf{Ensemble} \\
    \midrule
    \multirow{5}{*}{GCN}          & \textit{uniform}      & .767          & .858              & .821              \\
                                    & LST                   & .777          & .864              & .811              \\
                                    & SSM                   & .765          & .855              & .830              \\
                                    & Haversine             & .729          & .793              & .777              \\
    \midrule
    GAT                           & -                     & .772          & .795              & .821              \\
    \midrule
    Fusion-GCN                        & \textit{all}          & .795          & .849              & .849              \\
    \midrule
    MAGAT (ours)                  & \textit{all}          & \textbf{.897} & \textbf{.898}     & \textbf{.905}     \\
    \bottomrule
    \end{tabular}
\end{center}
\end{table}
\subsection{Limitations and future work}
Our evaluation focuses on the Italian peninsula, which occupies around $300.192$ km\textsuperscript{2} in the middle of Mediterranean Sea and it is characterized by high climatic variability and diversity. Concerning the West Nile virus, since 2002 the Italian Ministry of Health has implemented a veterinary surveillance plan to monitor the viral introduction and circulation of WNV in the whole country. The virus circulation detected in animals (hereafter veterinary cases) is mandatory registered by the local veterinary authorities into the National Animal Disease Notification System (SIMAN) and forms the ground-truth database used in the analyses~\ref{sec:westnile}. Thus, our analysis takes into account a wide geographical area, including many eco-climatic conditions over a long period of time.

Concerning other diseases, our model is based on the eco-climatic parameters which significantly influence the vectors population density and the amplification of viral transmission. These modelled conditions can be used for the study of other vector-borne diseases (\textit{e.g.}, Usutu), but are not suitable for the evaluation of diseases that operate via a directly transmitted infection (\textit{e.g.}, Sars-Cov-2 infection).

Finally, our proposal could be applied to other geographical areas to model the WNV circulation. However, it is necessary that these have the environmental and climatic conditions that allow the spread of the vector species and hosts.
\section{Conclusion}
This work upholds the benefits of multi-band satellite images for shaping the circulation of the West Nile Virus. We built upon the widely-acknowledged relation between virus presence and the suitability of the surrounding environment to host specific vectors. Our inference schema goes beyond the only consideration of the climatic and ecological cues present in the geographic location of interest; we also attend to its close neighbouring locations and model interactions through a Graph Neural Network. Experimental results and several ablative studies reward our intuition, providing further insights regarding the relevance of circulation and seasonality modeling. In future works, we plan to extend our approach to embrace even multi-horizon temporal dynamics.



\section*{Acknowledgment}
This work was done within the UNIMORE project ‘AI4V’. Funding was provided by the Italian Ministry of Health -- www.salute.gov.it -- (IZSAM 01/18 RC, Current Research 2018 Artificial intelligence and remote sensing: innovative methods for monitoring the vectors and the associated ecological/environmental variables) and by ESA EO Science for Society Permanently Open Call for Proposals EOEP-5 BLOCK 4 (ESA AO/1-9101/17/I-NB), (ESA Contract No.\ 4000128146/19/I-DT, project 'AIDEO' -- AI and EO as Innovative Methods for Monitoring West Nile Virus Spread). The view expressed in this paper can in no way be taken to reflect the opinion of the European Space Agency. Finally, the authors thank Progressive Systems Srl for their valuable support, in particular for the preparation and provision of EO Analysis Ready Datasets in accordance with the requirements defined by AImageLab and IZSAM.
\newpage

\bibliographystyle{IEEEtran}
\bibliography{IEEEabrv,main}

%

\vspace{-1.1cm}
\begin{IEEEbiography}[{\includegraphics[width=1in,height=1.25in,clip,keepaspectratio]{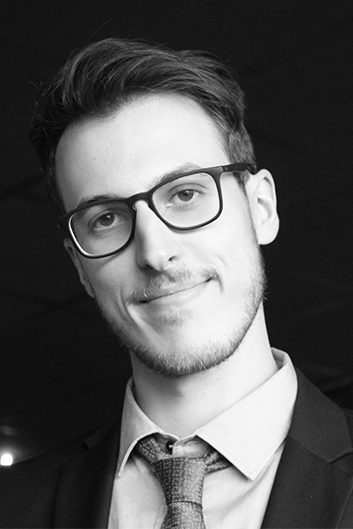}}]{Lorenzo Bonicelli}
is currently pursuing a Ph.D.\ degree at the University of Modena and Reggio Emilia, Italy, after receiving a bachelor's degree and a master's degree at the same university in 2018 and 2020 respectively. His current and past research interests include machine learning, deep learning, and the recent advances in continual learning and geometric deep learning.
\end{IEEEbiography}

\begin{IEEEbiography}[{\includegraphics[width=1in,height=1.25in,clip,keepaspectratio]{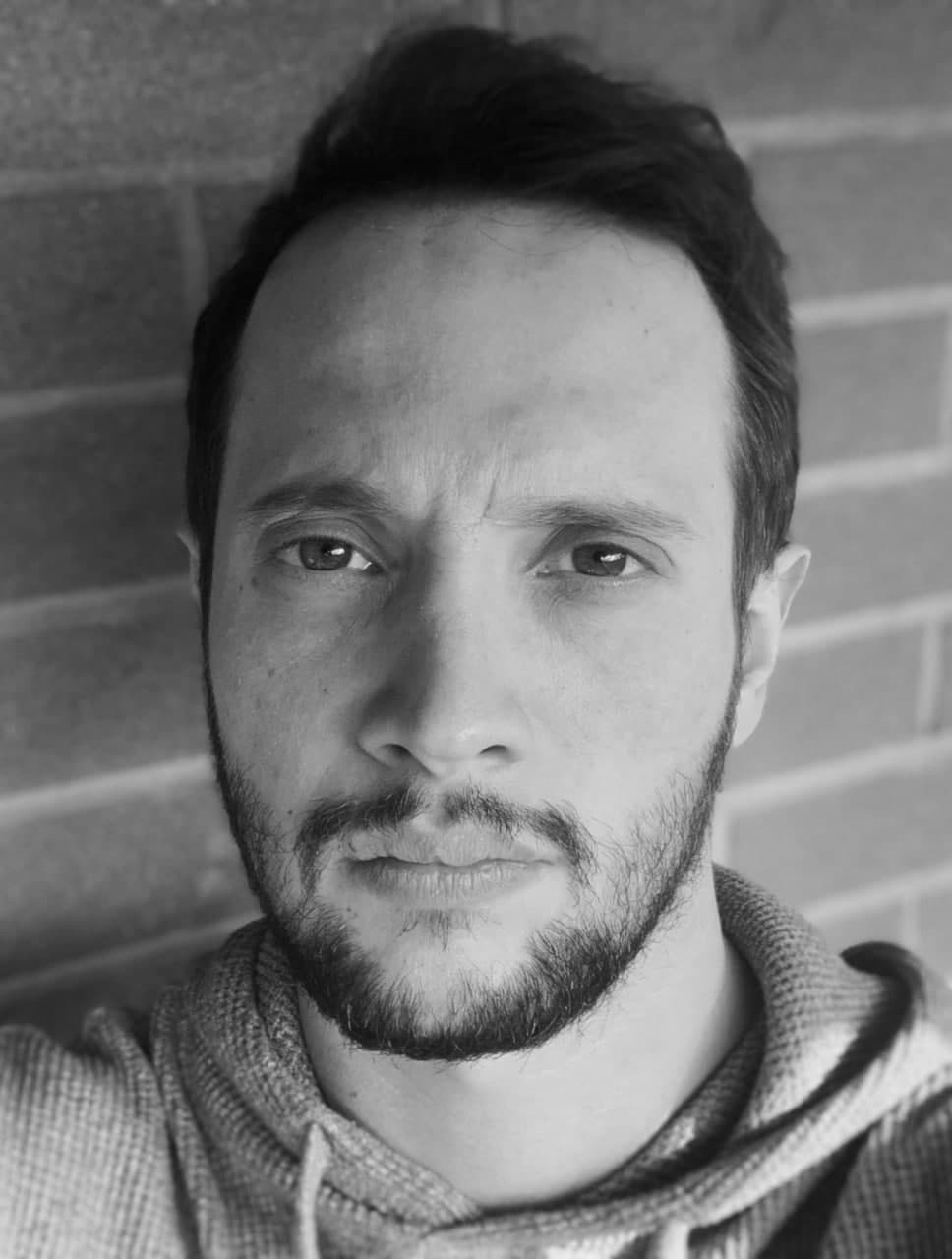}}]{Angelo Porrello}
obtained a Master's Degree in Computer Engineering in 2017 from the University of Modena and Reggio Emilia. He pursued a Ph.D.\ programme in ICT in the three-year period 2019-2021; currently, he is a Research Fellow within the AImageLab Group at the Department of Engineering “Enzo Ferrari”. His research interests focus on Deep Learning techniques: more precisely on Continual Learning, Re-Identification, and Anomaly Detection.
\end{IEEEbiography}

\begin{IEEEbiography}[{\includegraphics[width=1in,height=1.25in,clip,keepaspectratio]{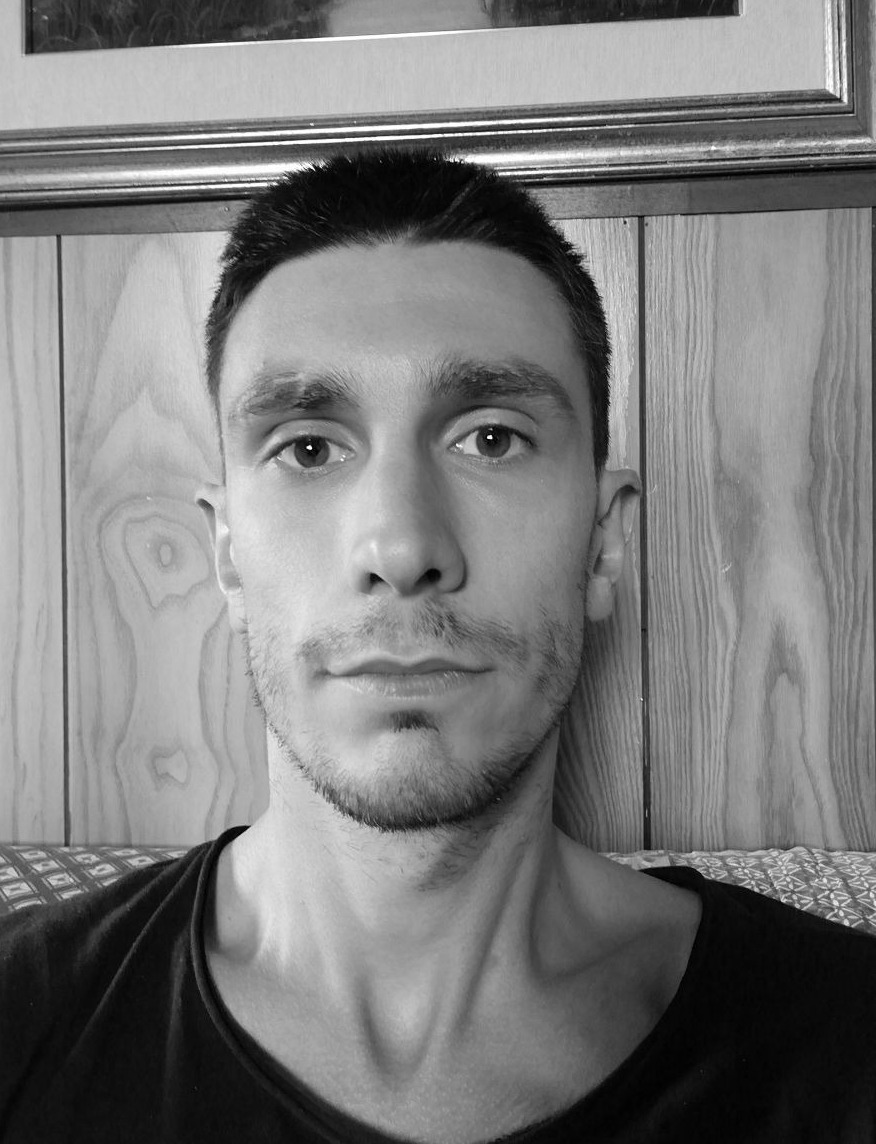}}]{Stefano Vincenzi}
is currently a machine learning engineer at Ammagamma. After receiving a Master's Degree in Computer Engineering in 2018 at the University of Modena and Reggio Emilia, he was a Research Fellow within the AImageLab Group. His research interests include machine learning and deep learning, with a focus on Image Segmentation, Text recognition, and Remote Sensing.
\end{IEEEbiography}

\begin{IEEEbiography}[{\includegraphics[width=1in,height=1.25in,clip,keepaspectratio]{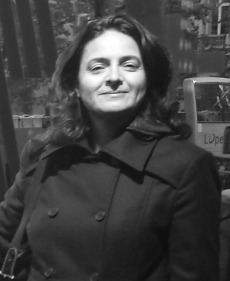}}]{Carla Ippoliti}
is a mathematician working at the National Reference Centre for Epidemiology in Italy. Her main interest is in spatial analysis of vectors and vector-borne diseases and the related driving factors. She has a great experience in Earth Observation data and their use in veterinary epidemiology.
\end{IEEEbiography}

\begin{IEEEbiography}[{\includegraphics[width=1in,height=1.25in,clip,keepaspectratio]{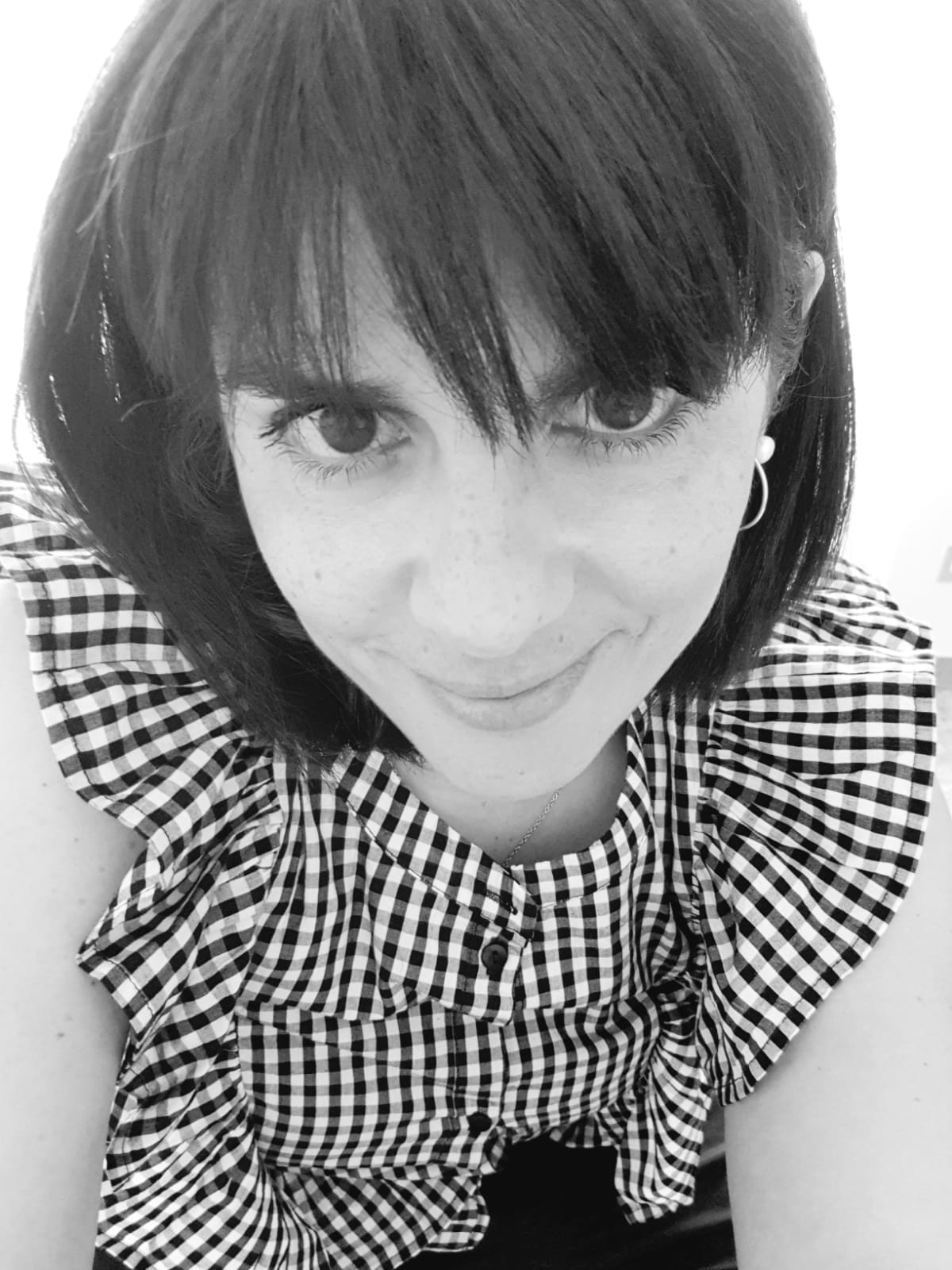}}]{Federica Iapaolo}
is a veterinarian working at the National Reference Centre for Foreign Animal Diseases (CESME) in Italy. Her main interest is West Nile virus (WNV) and Usutu Virus (USUV) epidemiology. She has experience in WNV surveillance and control activities in Italy. To date she is involved in the management of data flow collected within WNV and USUV and SIMAN (National Information System on animal disease) implementation to ensure the information system arrangement to the information debt towards WOAH, European Commission and EFSA.
\end{IEEEbiography}

\begin{IEEEbiography}[{\includegraphics[width=1in,height=1.25in,clip,keepaspectratio]{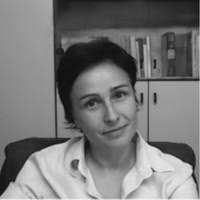}}]{Annamaria Conte}
Head of Statistics and GIS Unit of the National Reference Centre for Epidemiology of the Istituto Zooprofilattico Sperimentale dell'Abruzzo e del Molise "G. Caporale", Teramo (Italy). Her research focuses on epidemiological analytical methods and spatial epidemiology of major animal infectious diseases, including zoonoses, and on the identification of factors influencing the spread and persistence of vector borne diseases.
\end{IEEEbiography}

\begin{IEEEbiography}[{\includegraphics[width=1in,height=1.25in,clip,keepaspectratio]{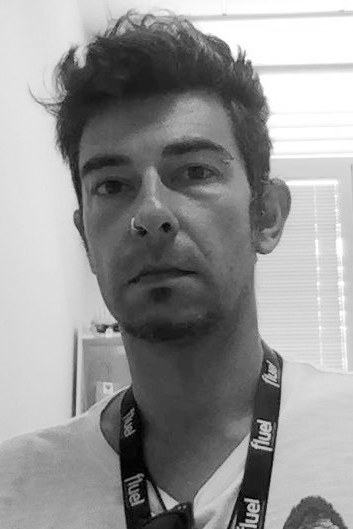}}]{Simone Calderara}
(Member, IEEE) received a computer engineering master's degree in 2005 and the Ph.D.\ degree in 2009 from the University of Modena and Reggio Emilia, where he is currently an assistant professor within the AImageLab group. His current research interests include computer vision and machine learning applied to human behavior analysis, visual tracking in crowded scenarios, and time series analysis for forensic applications. He is a member of the IEEE.
\end{IEEEbiography}




\end{document}